\documentclass[10pt,twocolumn,letterpaper]{article}

\usepackage{iccv}
\usepackage{times}
\usepackage{epsfig}
\usepackage{graphicx}
\usepackage{float} 
\usepackage{subfigure}
\usepackage{amsmath}
\usepackage{amssymb}

\usepackage{subfigure}
\usepackage{booktabs}
\usepackage{ textcomp }
\usepackage{color}
\usepackage{makecell}
\usepackage{multirow}
\usepackage{algorithmic}
\usepackage{cite}
\usepackage{authblk}

\usepackage{enumerate}

\usepackage[breaklinks=true,bookmarks=false]{hyperref}

\iccvfinalcopy 

\def\iccvPaperID{3439} 

\ificcvfinal\fi

\makeatletter
\newcommand{\printfnsymbol}[1]{
  \textsuperscript{\@fnsymbol{#1}}%
}
\makeatother

\begin{document}
\title{CFSNet: Toward a Controllable Feature Space for Image Restoration}

\makeatletter 
\renewcommand\AB@affilsepx{, \protect\Affilfont} 
\makeatother

\author[1]{Wei Wang \thanks{equal contribution}}
\author[2]{Ruiming Guo \printfnsymbol{1}}
\author[3]{Yapeng Tian}
\author[1]{Wenming Yang \thanks{corresponding author}}

 \affil[1]{Tsinghua University} 
 \affil[2]{Chinese University of Hong Kong} 
 \affil[3]{University of Rochester}
\renewcommand\Authands{ and } 
\date{} 

\twocolumn[{%
\renewcommand\twocolumn[1][]{#1}%
\maketitle
\begin{figure}[H]
\hsize=\textwidth 
\centering
\includegraphics[scale=0.51]{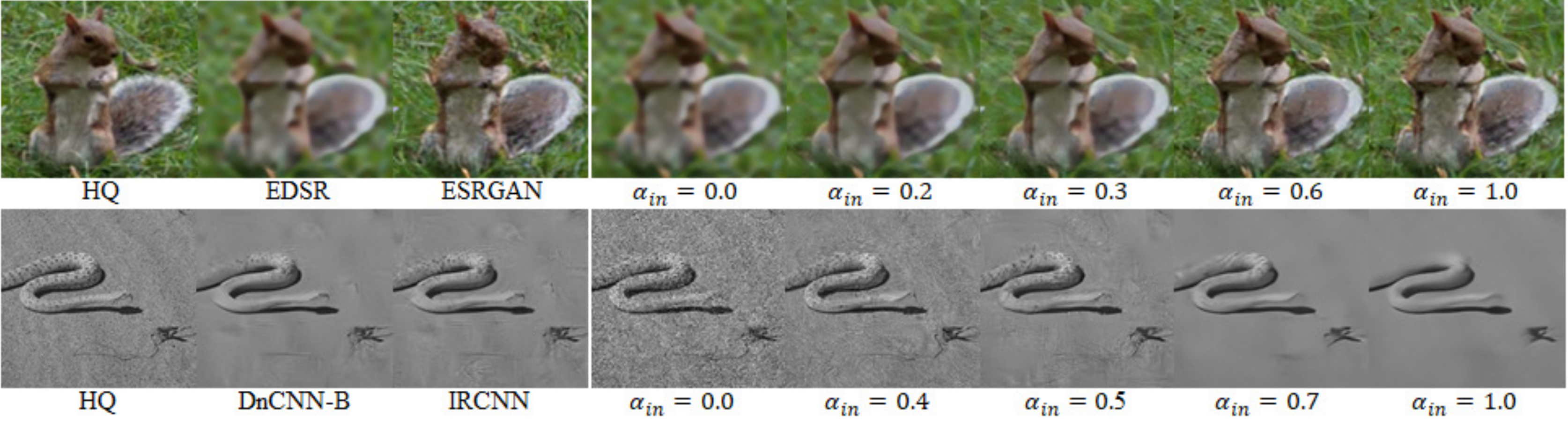}
\caption{The perception-distortion trade-off of image super-resolution (1st row) and the trade-off between noise reduction and detail preservation (2nd row). At test-time, when using CFSNet, users can easily adjust an input control variable $\alpha_{in}$ to attain the most satisfactory result according to personal preferences. In contrast, the fixed methods (\emph{e.g.}, EDSR, DnCNN-B) can not always guarantee the optimal visual quality.}
\label{first_visual}
\end{figure}
}]

\ificcvfinal\thispagestyle{empty}\fi

\renewcommand{\thefootnote}%
{\fnsymbol{footnote}}
\footnotetext[1]{Denotes equal contribution.}
\footnotetext[2]{Denotes corresponding author.}

\begin{abstract}
Deep learning methods have witnessed the great progress in image restoration with specific metrics (\emph{e.g.}, PSNR, SSIM). However, the perceptual quality of the restored image is relatively subjective, and it is necessary for users to control the reconstruction result according to personal preferences or image characteristics, which cannot be done using existing deterministic networks. This motivates us to exquisitely design a unified interactive framework for general image restoration tasks. Under this framework, users can control continuous transition of different objectives, \emph{e.g.}, the perception-distortion trade-off of image super-resolution, the trade-off between noise reduction and detail preservation. We achieve this goal by controlling the latent features of the designed network. To be specific, our proposed framework, named Controllable Feature Space Network (CFSNet), is entangled by two branches based on different objectives. Our framework can adaptively learn the coupling coefficients of different layers and channels, which provides finer control of the restored image quality. Experiments on several typical image restoration tasks fully validate the effective benefits of the proposed method. Code is available at \url{https://github.com/qibao77/CFSNet}.
\end{abstract}

\section{Introduction}
Image restoration is a classic ill-posed inverse problem that aims to recover high-quality images from damaged images affected by various kinds of degradations. According to the types of degradation, it can be categorized into different subtasks such as image super-resolution, image denoising, JPEG image deblocking, etc. 

The rise of deep learning has greatly facilitated the development of these subtasks. But these methods are often goal-specific, and we need to retrain the network when we deal with images different from the training dataset. Furthermore, most methods usually aim to pursue high reconstruction accuracy in terms of PSNR or SSIM. However, image quality assessment from personal opinion is relatively subjective, and low reconstruction distortion is not always consistent with high visual quality \cite{blau2018perception}. In addition, in many practical applications (\emph{e.g.}, mobile), it is often challenging to obtain user\textquotesingle s preference and the real degradation level of the corrupted images. All of these appeal to an interactive image restoration framework which can be applied to a wide variety of subtasks. However, to the best of our knowledge, currently there are few available networks which can satisfy both interactivity and generality requirements.

Some designs have been proposed to improve the flexibility of deep methods. Take image denoising for example, data augmentation is widely used to improve the generalization of a model. Training with the dataset which contains a series of noise levels, a single model can be applied to blind denoising task \cite{zhang2017beyond}. 
However, this method still produces a fixed reconstruction result of the input, which does not necessarily guarantee satisfactory perceptual quality (as shown in Fig. \ref{first_visual}). An alternative choice, Zhang \emph{et~al.} \cite{zhang2018ffdnet} concatenated a tunable noise level map with degraded images as input to handle blind image denoising task. Though this scheme is also user-friendly, it can not be generalized to other tasks. In image super-resolution, \cite{michelini2018multi} added noise to the input to control the compromise between perceptual quality and distortion. However, this scheme is specific for image super-resolution and cannot guarantee smooth and continuous control. 

In this paper, to rectify these weaknesses, we propose a novel framework equipped with controllability for human perception-oriented interactive image restoration. To be more specific, we realize the interactive control of the reconstruction result by tuning the features of each unit block, called coupling module. Each coupling module consists of a main block and a tuning block. The parameters of two blocks are obtained under two endpoint optimization objectives. Taking image super-resolution as an example, the main block is optimized for low distortion while the tuning block is optimized for high perceptual quality. Besides, as a key to achieving fine feature control, we assign the high-degree-of-freedom coupling coefficients adaptively learned from a control scalar to each coupling module.

Our main contributions can be summarized as follows:
\begin{itemize}

\item[$\blacktriangleright$] We propose a novel controllable end-to-end framework for interactive image restoration in a fine-grained way.
\item[$\blacktriangleright$] We propose a coupling module and an adaptive learning strategy of coupling coefficients to improve reconstruction performance.
\item[$\blacktriangleright$] Our CFSNet outperforms the state-of-the-art methods on super-resolution, JPEG image deblocking and image denoising in terms of flexibility and visual quality.

\end{itemize}
\section{Related Work}

\begin{bfseries} 
Image Restoration.
\end{bfseries}
Deep learning methods have been widely used in image restoration. \cite{kim2016accurate, tai2017memnet, tai2017image, lim2017enhanced, zhang2018residual, zhang2018image, yang2019lightweight} continuously deepen, widen or lighten the network structure, aiming at improving the super-resolution accuracy as much as possible. While \cite{johnson2016perceptual, ledig2017photo, sajjadi2017enhancenet, mechrez2018maintaining} paid more attention to the design of loss function to improve visual quality. Besides, \cite{blau2018perception, wang2018esrgan, liu2018multi} explored the perception-distortion trade-off. In \cite{dong2015compression}, Dong \emph{et~al.} adopted ARCNN built with several stacked convolutional layers for JPEG image deblocking. Zhang \emph{et~al.} \cite{zhang2018ffdnet} proposed FFDNet to make image denoising more flexible and effective. Guo \emph{et~al.} \cite{guo2019toward} designed CBDNet to handle blind denoising of real images. Different from these task-specific methods, \cite{zhang2017beyond, zhang2017learning, liu2018multi, liu2018non} proposed some unified schemes that can be employed to different image restoration tasks. However, these fixed networks are not flexible enough to deal with volatile user needs and application requirements.

 \begin{bfseries} 
Controllable Image Transformation.
\end{bfseries}
In high-level vision task, many technologies have been explored to implement controllable image transformation. \cite{lu2018attribute} and \cite{yu2018super} incorporated facial attribute vector into network to control facial appearance (\emph{e.g.}, gender, age, beard). In \cite{upchurch2017deep}, deep feature interpolation was adopt to implement automatic high-resolution image transformation. \cite{karras2019style} also proposed a scheme that controls adaptive instance normalization (AdaIN) in feature space to adjust high-level attributes. Shoshan \emph{et~al.} \cite{shoshan2018dynamic} inserted some tuning blocks in the main network to allow modification of the network. However, all of these methods are designed for high-level vision tasks and can not be directly applied to image restoration. To apply controllable image transformation to low-level vision tasks, Wang \emph{et~al.} \cite{wang2019deep} performed interpolation in the parameter space, but this method can not guarantee the optimality of the outputs, which inspires us to further explore fine-grain control of image restoration.
\section{Proposed Method}

\begin{figure*}[t]
\begin{center}
 \includegraphics[scale=0.45]{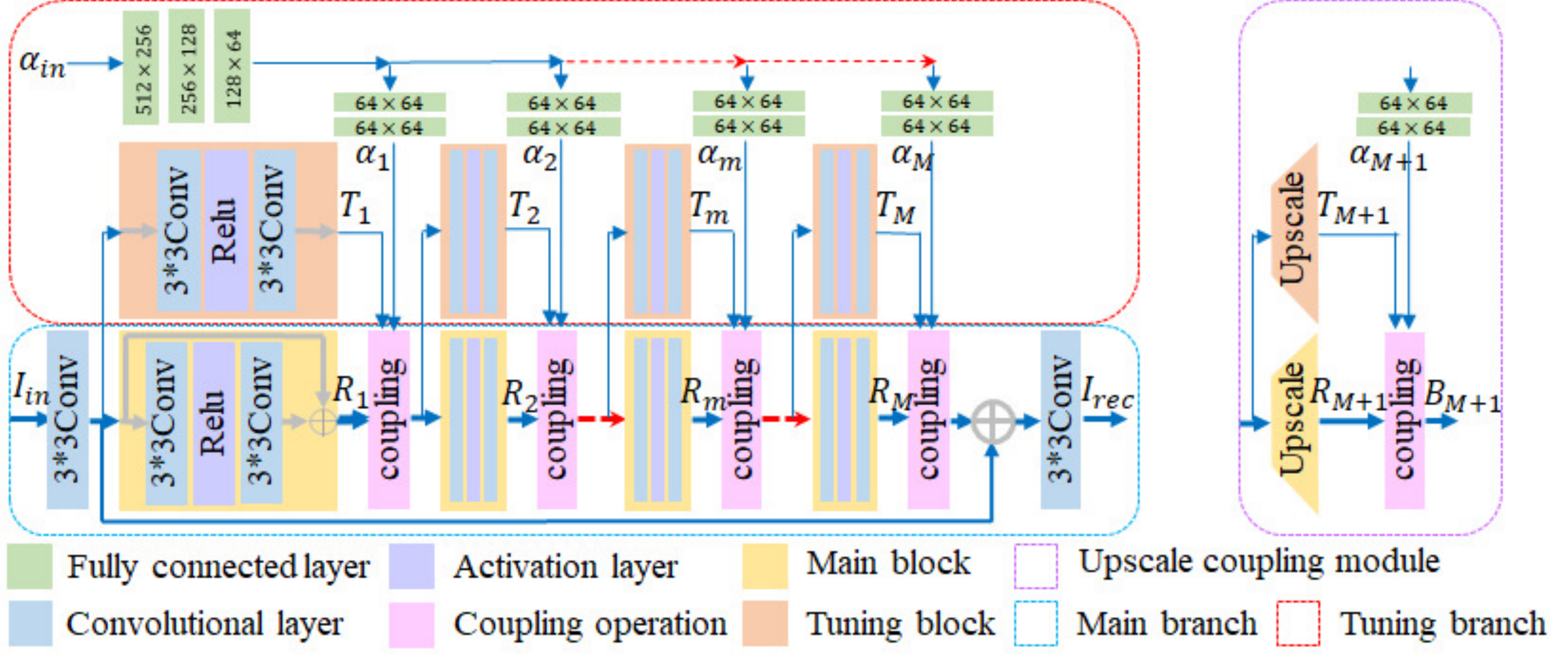}
\end{center}
\caption{The framework of our proposed controllable feature space network (CFSNet).}
\label{framework}
\end{figure*}


In this section, we first provide an overview of the proposed framework, called CFSNet and then, present the modeling process inspired by the image super-resolution problem. Instead of specializing for the specific super-resolution task, we finally generalize our CFSNet to multiple image restoration tasks, including denoising and deblocking. Moreover, we give the explicit model interpretation based on the manifold learning to show the intrinsic rationality of the proposed network. At the end of this section, we show the superiority and improvements of the proposed CFSNet through the detailed comparison with the current typical related methods.
\subsection{Basic Network Architecture}
As shown in Fig. \ref{framework}, our CFSNet consists of a main branch and a tuning branch. The main branch contains $M$ main blocks (residual blocks \cite{lim2017enhanced}) while the tuning branch contains $M$ tuning blocks with additional $2*M+3$ fully connected layers. A pair of main block and tuning block constitute a coupling module by using a coupling operation to combine the features of the two branches effectively. We take the original degraded image $I_{in}$ and the control scalar $\alpha_{in}$ as the input and output the restored image $I_{rec}$ as the final result.

In the beginning, we first use a $3\times 3$ convolutional layer to extract features from the degraded image $I_{in}$,
\begin{equation}
B_{0}=F_{in}(I_{in}),
\end{equation}
where $F_{in}(\cdot)$ represents the feature extraction function and $B_{0}$ serves as the input of next stage. Here, together with the input image $I_{in}$, we introduce a control scalar $\alpha_{in}$ to balance the different optimization goals. To be more specific, there are 3 shared fully connected layers to transform the input scalar $\alpha_{in}$ into multi-channels vectors and 2 independent fully connected layers to learn the optimal coupling coefficient for each coupling module:
\begin{equation}
\alpha_{m}=F_{m}^{ind}(F_{m}^{sha}(\alpha_{in})),
\label{alpha_mapping}
\end{equation}
where both $F_{m}^{sha}(\cdot)$ and $F_{m}^{ind}(\cdot)$ denote the function of the shared and independent fully connected layers, and $\alpha_{m}$ is the coupling coefficient vector of the $m$-th coupling module. Each coupling module couples the output of a main block and a tuning block as follows:
\begin{equation}
\begin{aligned}
B_{m}&=F_{m}(R_{m}, T_{m})\\
&=F_{m}(F_{m}^{main}(B_{m-1}), F_{m}^{tun}(B_{m-1})),
\end{aligned}
\end{equation}
where $F_{m}(\cdot)$ represents the $m$-th coupling operation, $R_{m}$ and $T_{m}$ denote the output features of $m$-th main block and $m$-th tuning block respectively, $F_{m}^{main}(\cdot)$ and $F_{m}^{tun}(\cdot)$ are the $m$-th main block function and $m$-th tuning block function respectively. To address the image super-resolution task, we add an extra coupling module consisting of the upscaling block before the last convolutional layer, as shown in Fig. \ref{framework}. Specifically, we utilize sub-pixel convolutional operation (convolution + pixel shuffle) \cite{shi2016real} to upscale feature maps. Finally, we use a $3\times 3$ convolutional layer to get the reconstructed image,
\begin{equation}
I_{rec}=F_{out}(B_{M}+B_{0}) \; or \;  I_{rec}=F_{out}(B_{M+1}),
\end{equation}
where $F_{out}(\cdot)$ denotes convolution operation. The overall reconstruction process can be expressed as
\begin{equation}
I_{rec}=F_{CFSN}(I_{in}, \alpha_{in};\theta_{main},\theta_{tun},\theta_{\alpha}),
\end{equation}
where $F_{CFSN}(\cdot)$ represents the function of our proposed CFSNet. $\theta_{main}$, $\theta_{tun}$ and $\theta_{\alpha}$ represent the parameters of main branch, all tuning blocks and all fully connected layers respectively. 

Since the two branches of our framework are based on different optimization objectives, in further detail, our training process can be divided into two steps:

\begin{enumerate}[\begin{bfseries} 
Step 1 
\end{bfseries}]

\item Set the control variable $\alpha_{in}$ as 0. Train the main branch with the loss function $L_{1}(I_{rec}, I_{g}; \theta_{main})$, where $I_{g}$ is the corresponding ground truth image.\label{Step1}

\item Set the control variable $\alpha_{in}$ as 1. Map the control variable $\alpha_{in}$ to different coupling coefficients $\left \{ \alpha_{m} \right \}$, fix parameters of the main branch and train the tuning branch with another loss function $L_{2}(I_{rec}, I_{g}; \theta_{tun},\theta_{\alpha})$.
\label{Step2}
\end{enumerate}

\subsection{Coupling module}
We now present the details of our coupling module. We mainly introduce our design from the perspective of image super-resolution. In order to balance the trade-off between perceptual quality and distortion, we usually realize it by modifying the penality parameter $\lambda$ of the loss terms \cite{blau2018perception},
\begin{equation}
L_{gen}=L_{distortion}+\lambda L_{adv},
\end{equation}
where $L_{distortion}$ denotes distortion loss (\emph{e.g.}, MSE and MAE), $L_{adv}$ contains GAN loss \cite{wang2018esrgan, gulrajani2017improved} and perceptual loss \cite{johnson2016perceptual}, $\lambda$ is a scalar. We usually pre-train the network with $L_{distortion}$ loss, then we fine-tune the network with combined loss $L_{gen}$ to reach a different working point for the trade-off determined by the value of $\lambda$. That is to say, if we regard pre-trained results as a reference point, then we can start from the reference point and gradually convert it to the result of another optimization goal.

However, it is not efficient to train a network for each different value of $\lambda$. In order to address this issue, we convert the control scalar $\lambda$ to the input and directly control the offset of the reference point in latent feature space. For this purpose, we implement a controllable coupling module to couple reference features learned with  $L_{distortion}$ and new features learned with $L_{gen}$ together. We set the feature based on distortion optimization as the reference point which is denoted as $R_{m}$. In the process of optimization based on perceptual quality, we keep the reference point unchanged and set $T_{m}-R_{m}$ as direction of change. In other words, in a coupling module, part of features are provided by reference information, and the other part are obtained from new exploration:
\begin{equation}
B_{m} = R_{m}+\alpha_{m}(T_{m}-R_{m})=(1-\alpha_{m})R_{m}+\alpha_{m}T_{m},
\label{coupling_module}
\end{equation}
where $T_{m}\in \mathbb{R}^{W\times H\times C}$, $R_{m}\in \mathbb{R}^{W\times H\times C}$, $\alpha_{m}\in \mathbb{R}^{C}$ denotes the $m$-th coefficient, and $C$ is the number of channels. 

It is worth noting that different main blocks provide different reference information, so we should treat them differently. We expect to endow each coupling module a different coupling coefficient to make full use of reference information. Therefore, our control coefficients $\left \{ \alpha_{m} \right \}$ are learned from optimization process. To be more specific, we use some fully connected layers to map a single input control scalar $\alpha_{in}$ into different coupling coefficients (Eq. \ref{alpha_mapping}). The proposed network will find the optimal coupling mode since our control coefficients $\left \{ \alpha_{m} \right \}$ are adaptive not fixed. 

Thanks to the coupling module and adaptive learning strategy, we can realize continuous and smooth transition by a single control variable $\alpha_{in}$. Moreover, if this framework achieves an excellent trade-off between perceptual quality and distortion for super-resolution, then can we generalize this model to other restoration tasks? After the theoretical analysis and experimental tests, we find that this framework is applicable to a wide variety of image restoration tasks. In the next section, we will provide a more general theoretical explanation of our model.

\subsection{Theoretical analysis}
Suppose there is a high-dimensional space containing all natural images. The degradation process of a natural image can be regarded as continuous in the space. So approximately, these degraded images are adjacent in the space. It is possible to approximate the reconstruction result of unknown degradation level with the results of known degradation levels. Unfortunately, natural images lie on an approximate non-linear manifold \cite{weinberger2006unsupervised}. As a result, simple image interpolation tends to introduce some ghosting artifacts or other unexpected details to final results. 

\begin{figure}[t]
\begin{center}
 \includegraphics[scale=0.6]{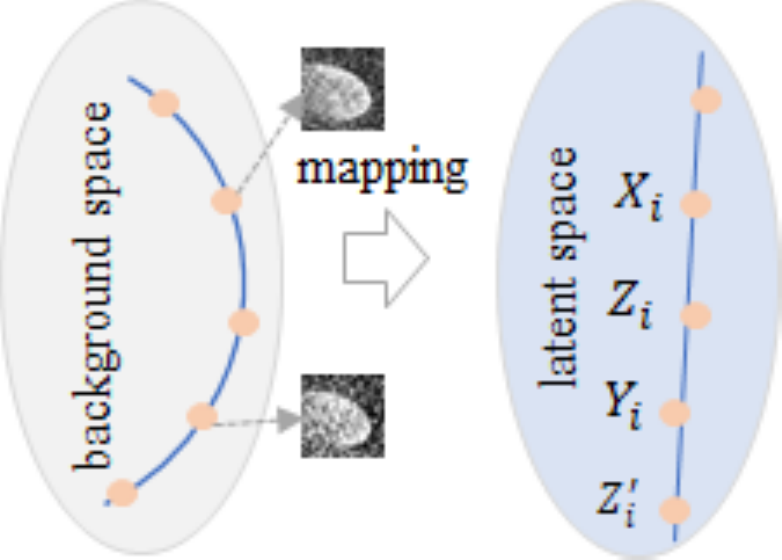}
\end{center}
\caption{Neural network mapping gradually disentangles data manifolds. We can represent unknown point with known point in latent space.}
\label{manifold_analysis}
\end{figure}

Instead of operating in the pixel space, we naturally turn our attention to the feature space. Some literatures indicate that the data manifold can be flattened by neural network mapping and we can approximate the mapped manifold as a Euclidean space \cite{bengio2013better, brahma2016deep, shoshan2018dynamic}. Based on this hypothesis, as shown in Fig. \ref{manifold_analysis}, we denote $X_{i}$ and $Y_{i}$ in the latent space as two endpoints respectively, and we can represent unknown point $Z_{i}$ as a affine combination of $X_{i}$ and $Y_{i}$: 
\begin{equation}
Z_{i} \approx \alpha_{i}X_{i}+(1-\alpha_{i})Y_{i},
\label{coupling_module2}
\end{equation}
where $\alpha_{i}$ is the $i$-th combination coefficient. This is exactly the formula of the controllable coupling module Eq. \ref{coupling_module}. However, we should also note that this hypothesis is influenced by the depth and width of CNN. In other words, we do not know the degree of flattening that can be achieved by different channels and different layers. Thus the combination coefficients of different channels and different layers should be discrepant. Besides, we hope to find the optimal combination coefficients through the optimization process:
\begin{equation}
\alpha^{*}=\mathop{\arg\min}_{\alpha_{ij}} \ \ \sum_{i=1}^{M}\sum_{j=1}^{C}\left [ Z_{ij}-\left ( \alpha_{ij}X_{ij}+\left ( 1-\alpha_{ij} \right )Y_{ij} \right ) \right ],
\label{combination_coefficients}
\end{equation}
where $\alpha^{*}=\left \{\alpha_{i,j}|i=1\cdots M,j=1\cdots C \right \}$ represents the optimal solution. However, it is difficult to directly obtain the optimal $\alpha^{*}$, because the unknown working point $Z_{ij}$ cannot in general be computed tractably. So we solve Eq. \ref{combination_coefficients} in an implicit way. Specifically, we map the input control variable $\alpha_{in}$ to different combination coefficients with some stacked fully connected layers, and then, we can approximate the above process into optimizing the parameters of the linear mapping network:
\begin{equation}
\alpha^{*}\approx \hat{\alpha}=F_{alpha}(\alpha_{in};\theta_{\alpha}),
\end{equation}
where $F_{alpha}$ denotes mapping function of $\alpha_{in}$, $\hat{\alpha}$ is the approximated solution of the optimal $\alpha^{*}$. Fortunately, this network can be embedded into our framework. Therefore, we can optimize the parameters of the linear mapping network and the tuning blocks in one shot. 
 The entire optimization process (corresponding to Step \ref{Step2}) can be expressed as
\begin{equation}
\theta_{tun},\theta_{\alpha}=\mathop{\arg\min}_{\theta_{tun},\theta_{\alpha}} \ \ L_{2}(F_{CFSN}(I_{in}, \alpha_{in};\theta_{tun},\theta_{\alpha}), I_{g}),
\label{whole_optimization}
\end{equation}

\label{section:Theoretical_analysis}

\subsection{Discussions}
\begin{figure}[!h]
\begin{center}
\includegraphics[width=0.9\linewidth]{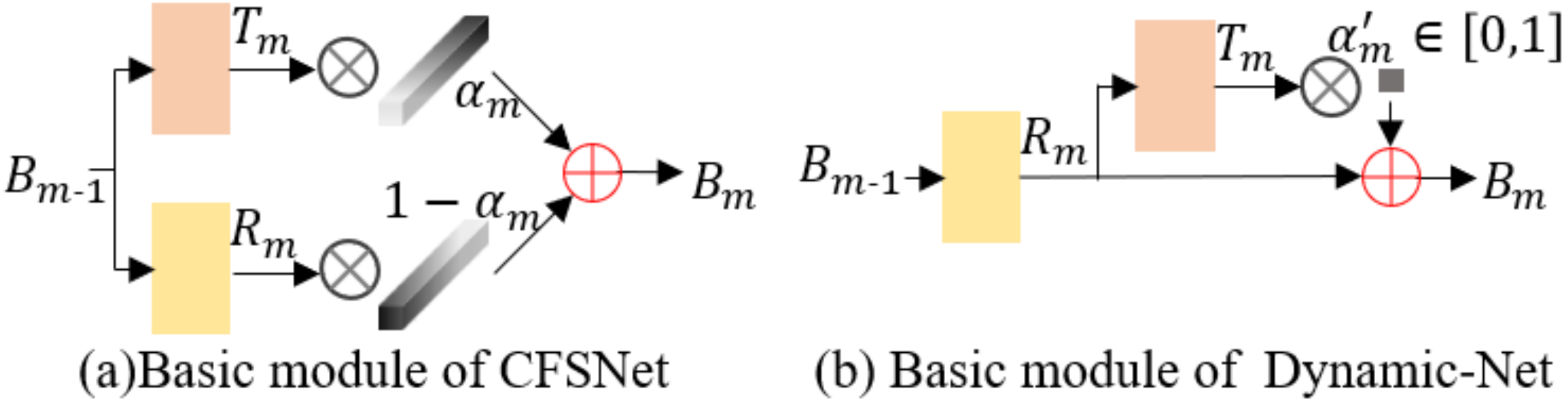}
\end{center}
\caption{Basic module. Yellow and orange bar represent main block and tuning block respectively.}
\label{basic_module}
\end{figure}

\begin{bfseries} 
Difference to Dynamic-Net
\end{bfseries}
 Recently, Dynamic-Net \cite{shoshan2018dynamic} realized interactive control of continuous image conversion. In contrast, there are two main differences between Dynamic-Net and our CFSNet. First of all, Dynamic-Net is mainly designed for image transformation tasks, like style transfer. It\textquotesingle s difficult to achieve the desirable results when we use Dynamic-Net directly for image restoration tasks. While motivated by super-resolution, we design the proposed CFSNet for low-level image restoration. Secondly, in Dynamic-Net, as shown in Fig. \ref{basic_module}, they directly tune multiple scalars $\left \{ \alpha_{m} \right \}$ to get different outputs. While, in our framework coupling coefficient $\alpha_{m}$ is a vector and is learned adaptively from the input scalar $\alpha_{in}$ through the optimization process. This is more user-friendly and we explain the reasonability of this design in Sec. \ref{section:Theoretical_analysis}.
 
\begin{bfseries} 
Difference to Deep Network Interpolation
\end{bfseries}
Deep Network Interpolation (DNI) is another choice to control the compromise between perceptual quality and distortion \cite{wang2018esrgan, wang2019deep}. DNI also can be applied to many low-level vision tasks \cite{wang2019deep}. However, this method needs to train two networks with the same architecture but different losses and will generate a third network to control. In contrast, our framework can achieve better interactive control with a unified end-to-end network. Moreover, our framework makes better use of reference information using coupling module. DNI performs interpolation in the parameter space to generate continuous transition effects, and interpolation coefficients are kept the same in the whole parameter space. However, this simple strategy can not guarantee the optimality of the outputs. While in our CFSNet, we perform the interpolation in the feature space and the continuous transition of the reconstruction effect is consistent with the variation of the control variable $\alpha_{in}$. We can produce a better approximation of the unknown working point. See Sec. \ref{section:super-resolution} for more experimental comparisons.

\section{Experiments}
In this section, we first demonstrate implementation details of our framework. Then we validate the control mechanism of our CFSNet. Finally, we apply our CFSNet to three classic tasks: image super-resolution, image denoising, JPEG image deblocking. All experiments validate the effectiveness of our model. Due to space limitations, more examples and analyses are provided in the appendix.
\subsection{Implement Details}
\begin{bfseries} 
\end{bfseries}
\begin{bfseries} 
\end{bfseries}
For image super-resolution task, our framework contains 30 main blocks and 30 tuning blocks (i.e., $M=30$). As for the other two tasks, the main branch parameters of our CFSNet are kept similar to that of the compared method \cite{zhang2017beyond} (i.e., $M=10$) for a fair comparison. Furthermore, we first generate a 512-dimensional vector with values of all 1. Then we multiply it by the control scalar $\alpha_{in}$ to produce a control input. All convolutional layers have 64 filters and the kernel size of each convolutional layer is $3\times 3$. We use the method in \cite{he2015delving} to perform weight initialization. For both training stage of all tasks, we use the ADAM optimizer \cite{kingma2014adam} by setting $\beta_{1}=0.9$, $\beta_{2}=0.999$, and $\varepsilon = 10^{-8}$ with the initial learning rate 1e-4. We adopt 128 as the minibatch size in image denoising task and set it as 16 in the other three tasks. We use PyTorch to implement our network and perform all experiments on GTX 1080Ti GPU. 

For image denoising and JPEG image deblocking, we follow the settings as in \cite{zhang2017beyond} and \cite{dong2015compression} respectively. The training loss function in Step \ref{Step1} and Step \ref{Step2} remains unchanged: $L_{1}(I_{rec}, I_{g})=L_{2}(I_{rec}, I_{g})$. In particular, for image denoising, we input the degraded images of noise level 25 when we train the main branch in Step \ref{Step1} and we input the degraded images of noise level 50 when we train the tuning branch in Step \ref{Step2}. Training images are cut into $40\times 40$ patches with a stride of 10. And the learning rate is reduced by 10 times every 50000 steps. For JPEG deblocking, we set quality factor as 10 in the first training stage and change it to 40 in the second training stage. Besides, we choose $48\times 48$ as patch size and the learning rate is divided by 10 every 100000 steps. For image super-resolution, we first train the main branch with objective MAE loss, then we train the tuning branch with objective $L_{2}=L_{mae}+0.01L_{gan}+0.01L_{per}$, where $L_{mae}$ denotes mean absolute error (MAE), $L_{gan}$ represents wgan-gp loss \cite{gulrajani2017improved} and $L_{per}$ is a variant of perceptual loss \cite{wang2018esrgan}. We set HR patch size as $128\times 128$ and we multiply the learning rate by 0.6 every 400000 steps.

\subsection{Ablation Study}
\begin{figure}[t]
\centering  
\subfigure[image denoising]{
\label{Ablation_alpha.sub.1}
\includegraphics[width=0.22\textwidth]{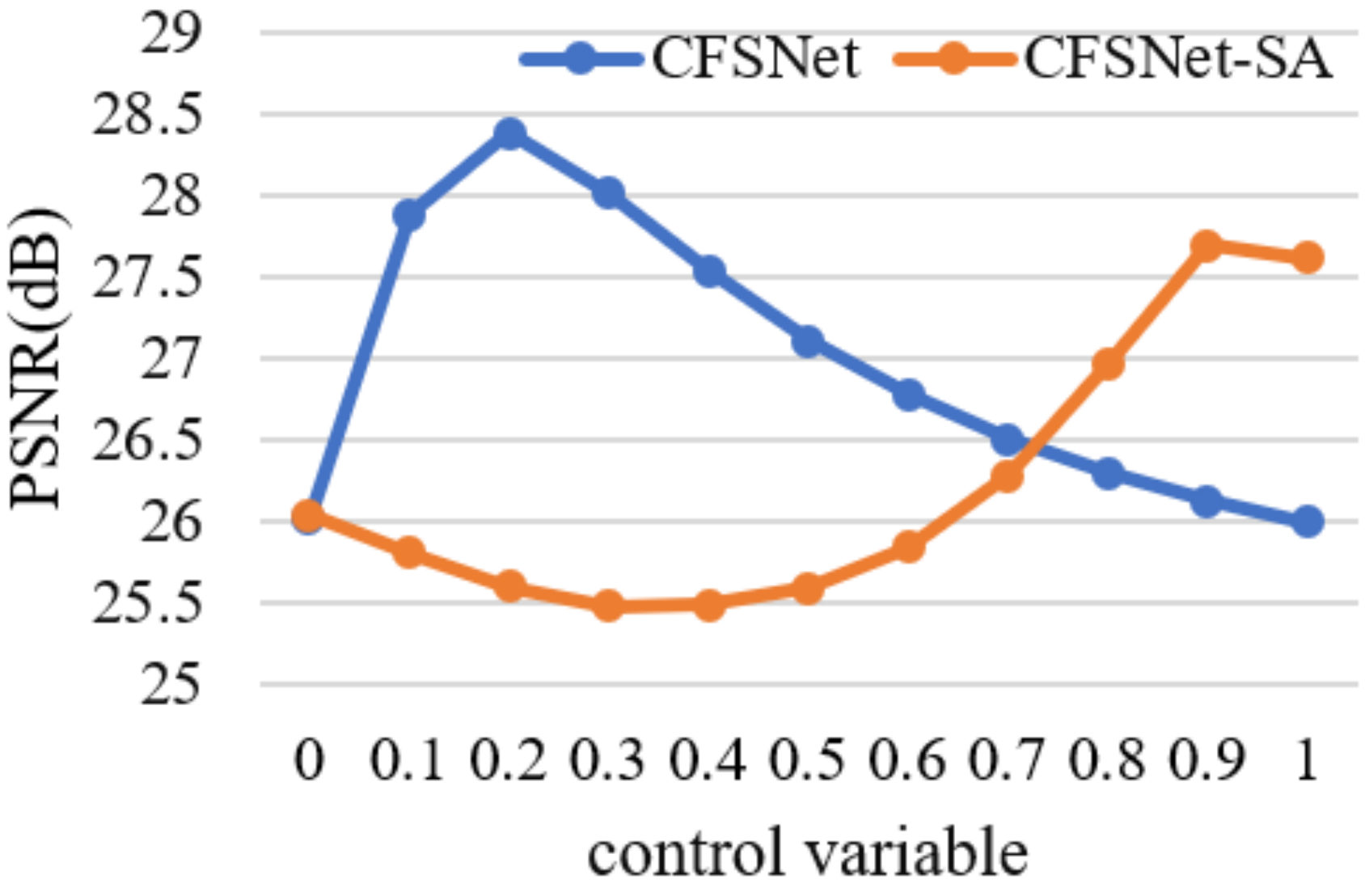}}
\subfigure[JPEG image deblocking]{
\label{Ablation_alpha.sub.2}
\includegraphics[width=0.22\textwidth]{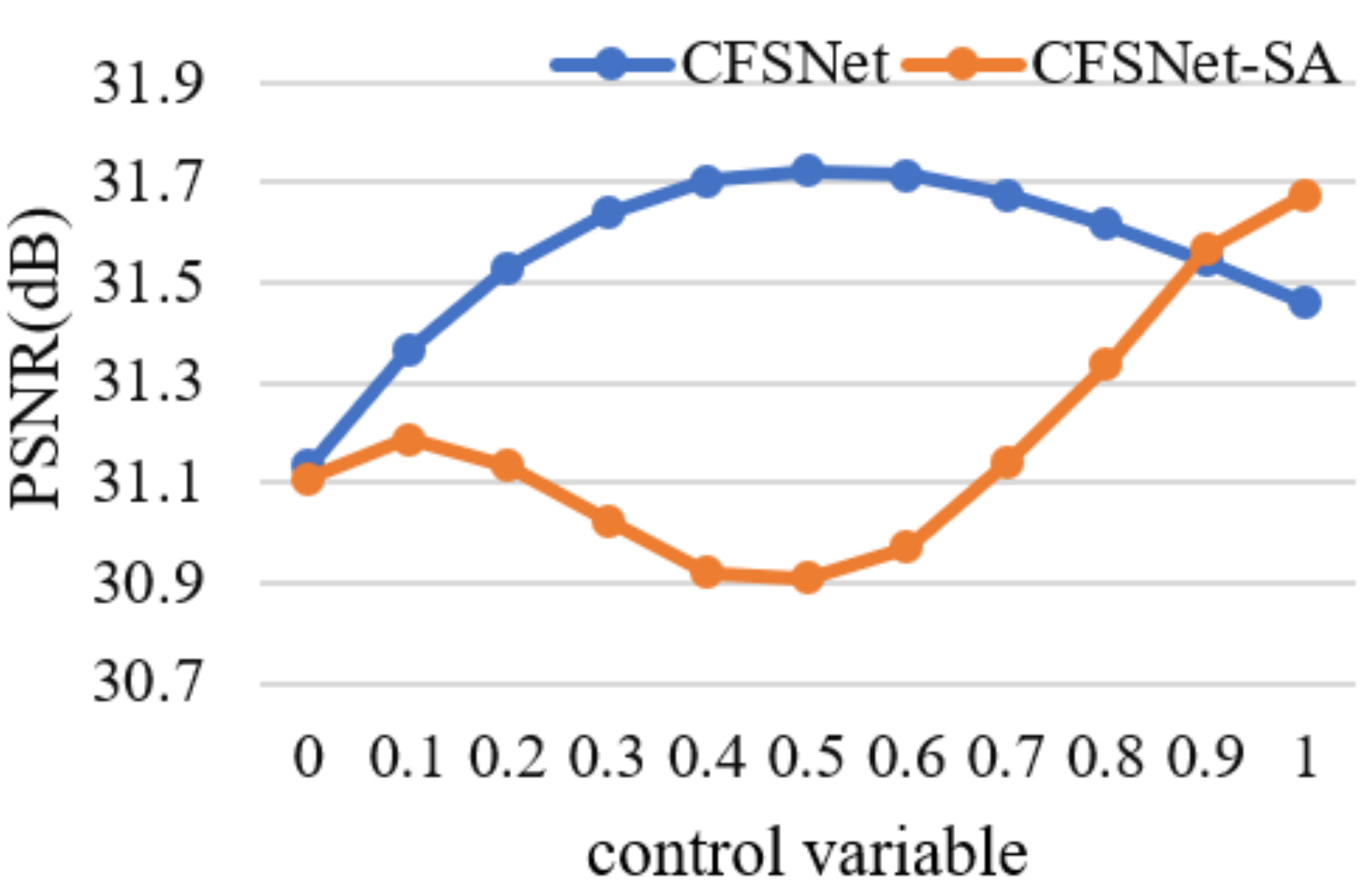}}
\caption{Average PSNR curve for noise level 30 on the BSD68 dataset and same curve for quality factor 20 on the LIVE1 dataset.}
\label{Ablation_alpha}
\end{figure}

Fig. \ref{Ablation_alpha} presents the ablation study on the effects of adaptive learning coupling coefficients strategy. We directly set the coupling coefficients of different channels and different layers as the same $\alpha_{in}$ in CFSNet-SA. That is, compared with CFSNet, CFSNet-SA removes the linear mapping network of control variable $\alpha_{in}$. Otherwise, we keep the training process of CFSNet-SA consistent with CFSNet. We can find that, no matter in denoising task or in deblocking task, the best restored result of CFSNet is better than that of CFSNet-SA for unseen degradation level. In particular, the curve of CFSNet is concave-shaped, which means that there is a bijective relationship between the reconstruction effect and the control variable. In contrast, there is no obvious change law in the curve of CFSNet-SA. The reason is that adaptive coupling coefficients help to produce better intermediate features. This merit provides more friendly interaction control. What\textquotesingle s more, JPEG deblocking task is more robust to control variable than image denoising task, we speculate that this is because JPEG images of different degradation levels are closer in the latent space.

\subsection{Image Super-resolution}
For image super-resolution, we adopt a widely used DIV2K training dataset \cite{agustsson2017ntire} that contains 800 images. We down-sample the high resolution image using MATLAB bicubic kernel with a scaling factor of 4. Following \cite{michelini2018multi, wang2018esrgan}, we evaluate our models on PIRM test dataset provided in the PIRM-SR Challenge \cite{blau20182018}. We use the perception index (PI) to measure perceptual quality and use RMSE to measure distortion. Similar to the PIRM-SR challenge, we choose EDSR \cite{lim2017enhanced}, CX \cite{mechrez2018maintaining} and EnhanceNet \cite{sajjadi2017enhancenet} as baseline methods. Furthermore, we also compare our CFSNet with another popular trade-off method, deep network interpolation \cite{wang2019deep, wang2018esrgan}. We directly use source code from ESRGAN \cite{wang2018esrgan} to produce SR results with different perceptual quality, namely ESRGAN-I.

\begin{figure}[!t]
\centering
\includegraphics[scale=0.5]{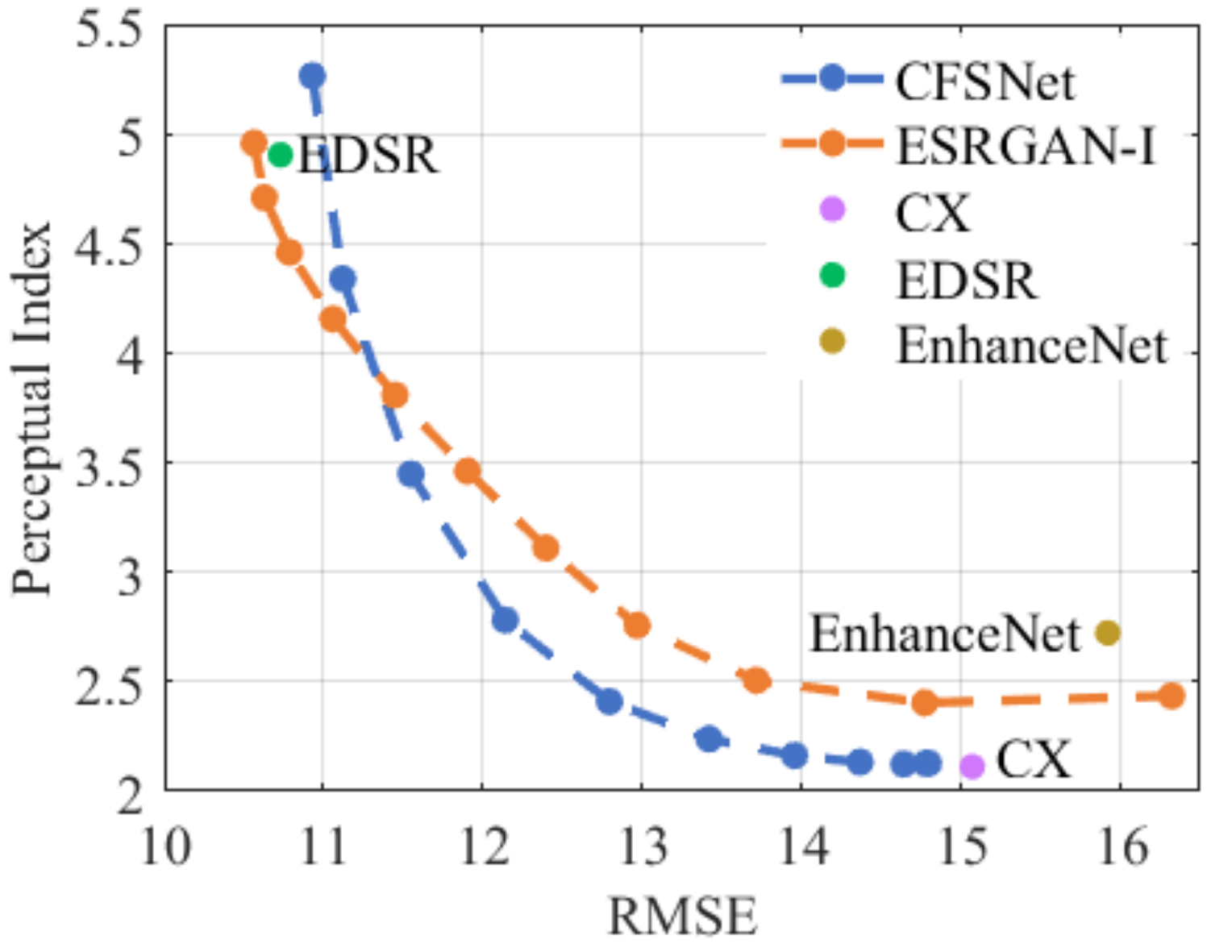}
\caption{Perception-distortion plane on PIRM test dataset. We gradually increase $\alpha_{in}$ from 0 to 1 to generate different results from distortion point to perception point.}
\label{Perception_distortion}
\end{figure}

\begin{figure*}[ht]
\begin{center}
 \includegraphics[scale=0.537]{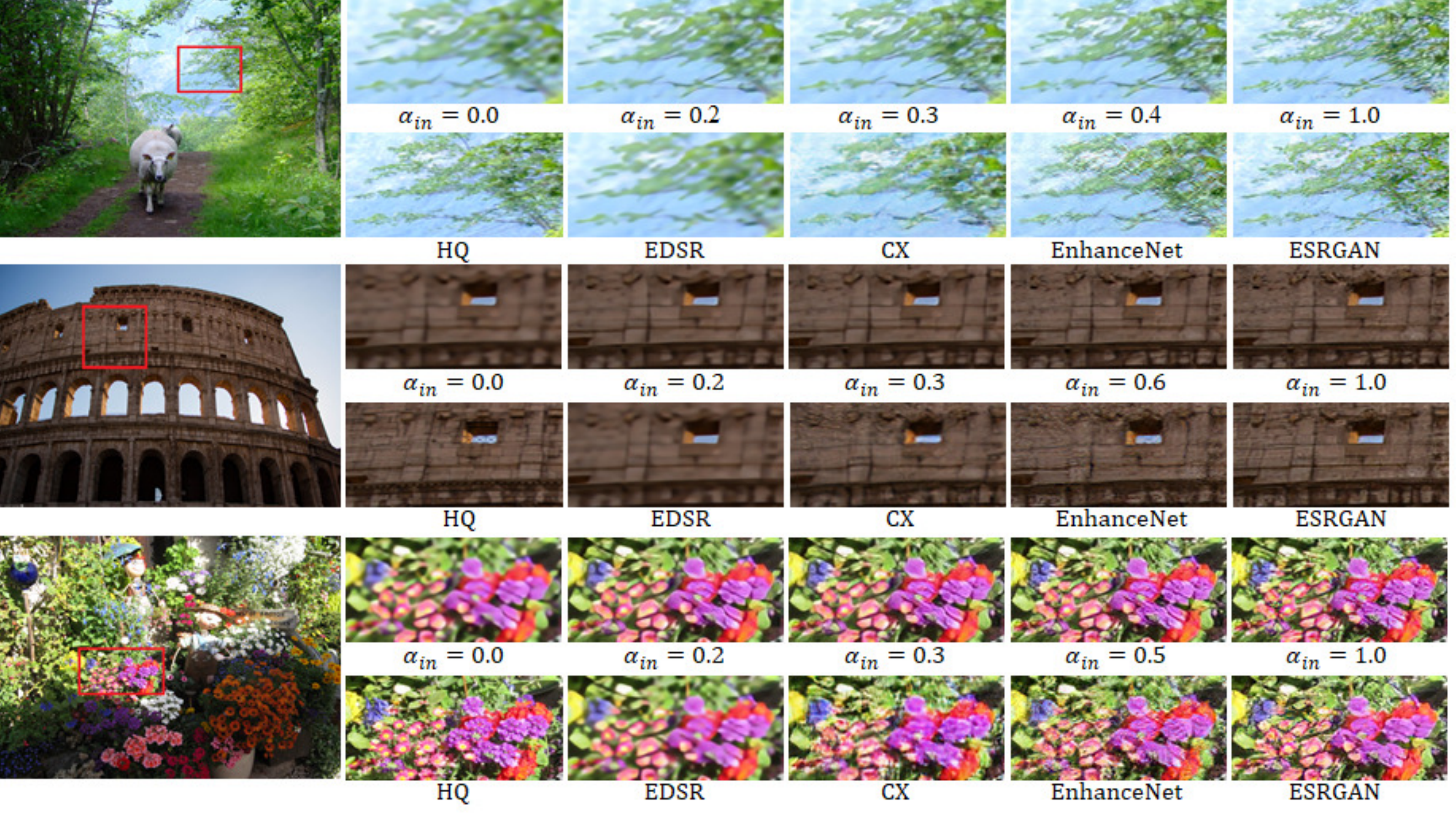}
\end{center}
\caption{Perceptual and distortion balance of ``215'', ``211'' and ``268'' (PIRM test dataset) for $4\times$ image super-resolution.}
\label{sr_compare}
\end{figure*}

\begin{figure*}[ht]
\centering
\includegraphics[scale=0.53]{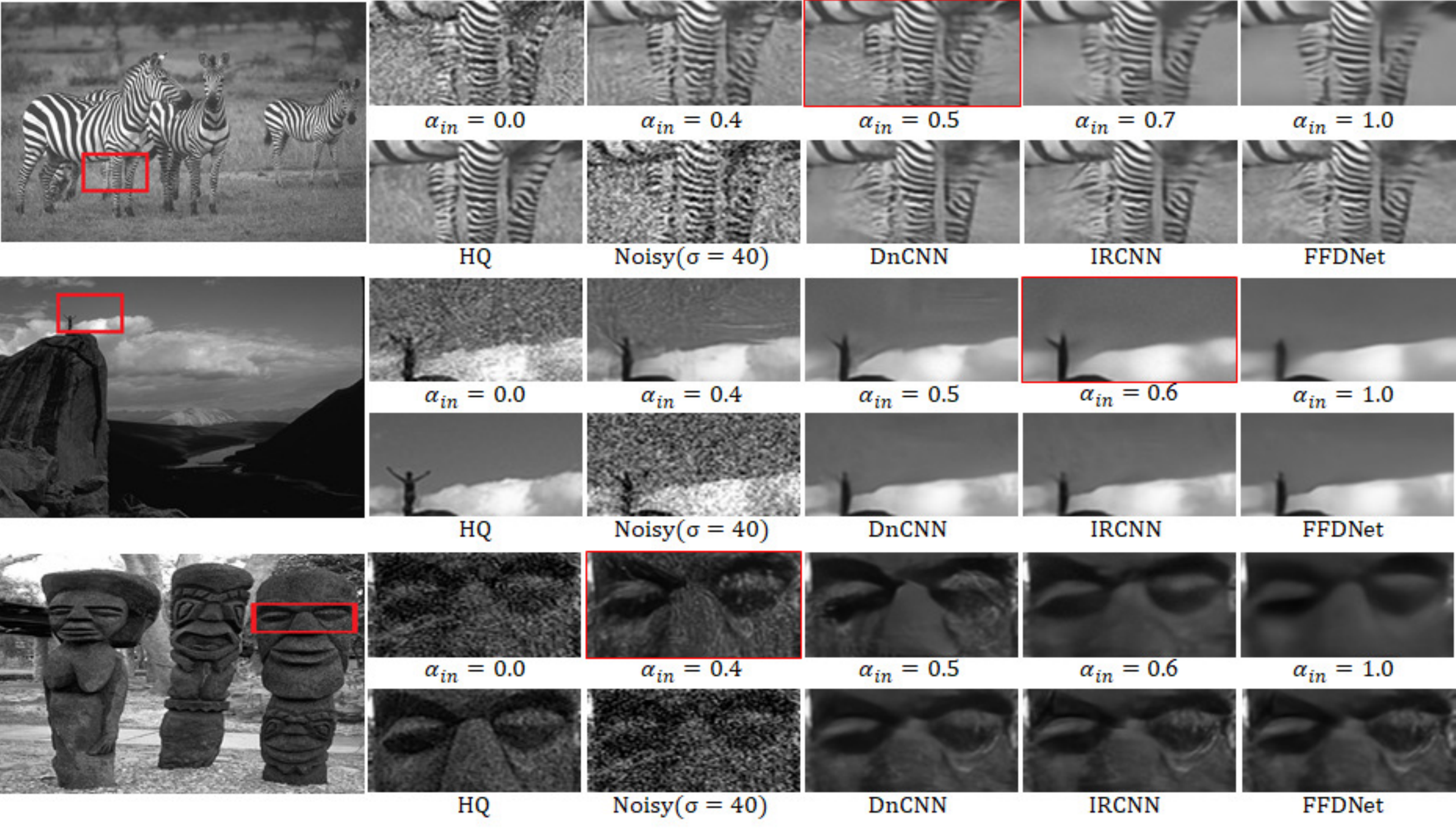}
\caption{Gray image denoising results of ``test051'' ``test017'' and ``test001'' (BSD68) with unknown noise level $\sigma=40$. $\alpha_{in}=0.5$ corresponds to the highest PSNR results, and the best visual results are marked with red boxes.}
\label{noise40_51and44}
\end{figure*}

\begin{figure*}[ht]
\centering
\includegraphics[width=1.0\textwidth, height=0.18\textheight]{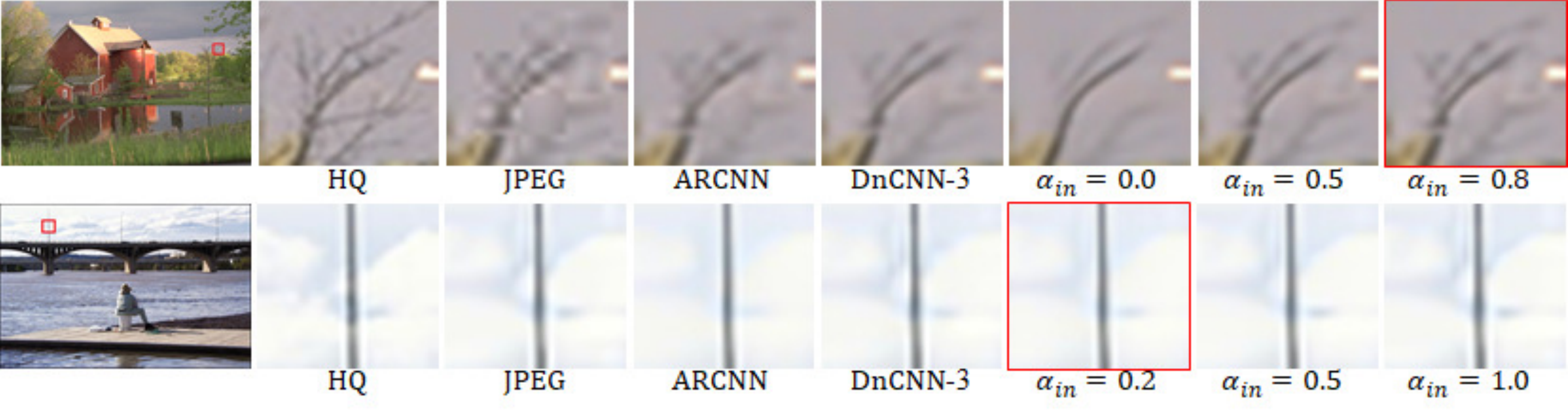}
\caption{JPEG image artifacts removal results of ``house'' and ``ocean'' (LIVE1) with unknown quality factor 20. $\alpha_{in}=0.5$ corresponds to the highest PSNR results, and the best visual results are marked with red boxes.}
\label{jpeg_20}
\end{figure*}
Fig. \ref{sr_compare} and Fig. \ref{first_visual} show the visual comparison between our results and the baselines. We can observe that CFSNet can achieve a mild transition from low distortion results to high perceptual quality results without unpleasant artifacts. In addition, it can be found that our CFSNet outperforms the baselines on edges and shapes. Due to different user preferences, it is necessary to allow users to adjust the reconstruction results freely.

We also provide quantitative comparisons on PIRM test dataset. Fig. \ref{Perception_distortion} shows the perception-distortion plane. As we can see, CFSNet improves the baseline (EnhanceNet) in both perceptual quality and reconstruction accuracy. The blue curve shows that our perception-distortion function is steeper than ESRGAN-I (orange curve). Meanwhile, CFSNet performs better than ESRGAN-I in most regions, although our network is lighter than ESRGAN. This means that our result is closer to the theoretical bound of perception-distortion. 


\label{section:super-resolution}

\subsection{Image Denoising}
In image denoising experiments, we follow \cite{zhang2017beyond} to use 400 images from the Berkeley Segmentation Dataset (BSD) \cite{martin2001database} as the training set. We test our model on BSD68 \cite{martin2001database} using the mean PSNR as the quantitative metric. Both training set and test set are converted to gray images. We generate the degraded images by adding Gaussian noise of different levels (\emph{e.g.}, 15, 25, 30, 40, and 50) to clean images. 


We provide visual comparison in Fig. \ref{noise40_51and44} and Fig. \ref{first_visual}. As we can see, users can easily control $\alpha_{in}$ to balance noise reduction and detail preservation. 
It is worth noting that, our highest PSNR results ($\alpha_{in}=0.5$) have similar visual quality with other methods, but it does not necessarily mean the best visual effects, for example, the sky patch of `test017'' enjoys a smoother result when $\alpha_{in}=0.6$. Users can personalize each picture and choose their favorite results by controlling $\alpha_{in}$ at test-time. 

In addition to perceptual comparisons, we also provide objective quantitative comparisons. We change $\alpha_{in}$ from 0 to 1 with an interval of 0.1 for preset noise range ($\sigma \in [25,30,40,50]$). Then we choose the final result according to the highest PSNR. We compare our CFSNet with several state-of-the-art denoising methods: BM3D \cite{4271520}, TNRD \cite{chen2017trainable}, DnCNN \cite{zhang2017beyond}, IRCNN \cite{zhang2017learning}, FFDNet \cite{zhang2018ffdnet}. More interestingly, as shown in Tab. \ref{Benchmark_denoise}, our CFSNet is comparable with FFDNet on the endpoint ($\sigma=25$ and $\sigma=50$), but our CFSNet still achieves the best performance on the noise level 30 which is not contained in the training process. Moreover, our CFSNet can even deal with unseen outlier ($\sigma=15$). This further verifies that we can obtain a good approximation of the unknown working point.

\begin{table}[!t]
\centering
\caption{Benchmark image denoising results. The average PSNR(dB) for various noise levels on (gray) BSD68. $^{*}$denotes unseen noise levels for our CFSNet in the training stage.}
\begin{tabular}{|c|c|c|c|c|}
\hline
methods&$\sigma=15^{*}$&$\sigma=25$&$\sigma=30^{*}$&$\sigma=50$\\
\hline
\hline
BM3D& 31.08&28.57 &27.76 & 25.62 \\
\hline
TNRD&31.42&28.92 & 27.66 & 25.97 \\
\hline
DnCNN-B&31.61&29.16 & 28.36 & 26.23 \\
\hline
IRCNN&31.63&29.15 & 28.26 & 26.19 \\
\hline
FFDNet&31.63&29.19 &28.39&26.29 \\
\hline
CFSNet&31.29&29.24&28.39&26.28 \\
\hline
\end{tabular}
\label{Benchmark_denoise}
\end{table}

\subsection{JPEG Image Deblocking}
We also apply our framework to reduce image compression artifacts. As in \cite{dong2015compression, zhang2017beyond, liu2018multi}, we adopt LIVE1 \cite{moorthy2009visual} as the test dataset and use the BSDS500 dataset \cite{martin2001database} as base training set. For a fair comparison, we perform training and evaluating both on the luminance component of the YCbCr color space. We use the MATLAB JPEG encoder to generate JPEG deblocking input with four JPEG quality settings q = 10, 20, 30, 40.

We select the deblocking result in the same way as the image denoising task. We select SA-DCT \cite{foi2007pointwise}, ARCNN \cite{dong2015compression}, TNRD \cite{chen2017trainable} and DnCNN \cite{zhang2017beyond} for comparisons. Tab. \ref{Benchmark_deblocking} shows the JPEG deblocking results on LIVE1. Our CFSNet achieves the best PSNR results on all compression quality factors. Especially, our CFSNet does not degrade too much and still achieves 0.12 dB and 0.18 dB improvements over DnCNN-3 on quality 20 and 30 respectively, although JPEG images of quality 20 and 30 never appear in training process. Fig. \ref{jpeg_20} shows visual results of different methods on LIVE1. Too small $\alpha_{in}$ produces too smooth results, while too large $\alpha_{in}$ leads to incomplete artifacts elimination. Compared to ARCNN \cite{dong2015compression} and DnCNN \cite{zhang2017beyond}, our CFSNet can make a better compromise between artifacts removal and details preservation.



\begin{table}[!t]
\centering
\caption{Benchmark JPEG deblocking results. The average PSNR(dB) on the LIVE1 dataset. $^{*}$ denotes unseen quality factors for our CFSNet in the training stage.}
\begin{tabular}{|c|c|c|c|c|}
\hline
methods&$q=10$&$q=20^{*}$&$q=30^{*}$&$q=40$\\
\hline
\hline
JPEG& 27.77&30.07 &31.41 & 32.35 \\
\hline
SA-DCT&28.65&30.81 & 32.08 & 32.99 \\
\hline
ARCNN&28.98&31.29 & 32.69 & 33.63 \\
\hline
TNRD&29.15&31.46 & 32.84 & N/A \\
\hline
DnCNN-3&29.19&31.59 &32.98&33.96 \\
\hline
CFSNet&29.36&31.71&33.16&34.16 \\
\hline
\end{tabular}
\label{Benchmark_deblocking}
\end{table}

\section{Conclusion}

In this paper, we introduce a new well-designed framework which equipped with flexible controllability for image restoration. The reconstruction results can be finely controlled using a single input variable with an adaptive learning strategy of coupling coefficients. Besides that, it is capable of producing high quality images on image restoration tasks such as image super-resolution, image blind denoising and image blind deblocking, and it outperforms the existing state-of-the-art methods in terms of user-control flexibility and visual quality. Future works will focus on the expansion of the multiple degraded image restoration tasks.

\begin{bfseries} 
\noindent Acknowledgements.
\end{bfseries}
This work was supported by the Natural Science Foundation of China (Nos. 61471216 and 61771276) and the Special Foundation for the Development of Strategic Emerging Industries of Shenzhen (Nos. JCYJ20170307153940960 and JCYJ20170817161845824).

{\small
\bibliographystyle{ieee}
\bibliography{arxiv_tex}
}

\clearpage

\appendix
\renewcommand{\appendixname}{Appendix~\Alph{section}}

\section*{Appendix}
\par We first provide more model analyses about the proposed coupling module in Sec. \ref{Model_Analyses}. Then we extend our model to two different image restoration tasks in Sec. \ref{Color_Image_Denoising} and Sec. \ref{SR_blur}, \emph{i.e.}, color image denoising and general image super-resolution. Furthermore, additional results can be found in Sec. \ref{Additional_Results}.

\section{More Examples and Analyses}

\begin{figure}[!hbt]
\centering  
\subfigure[image denoising]{
\label{Ablation_alpha.sub.1}
\includegraphics[width=0.45\textwidth]{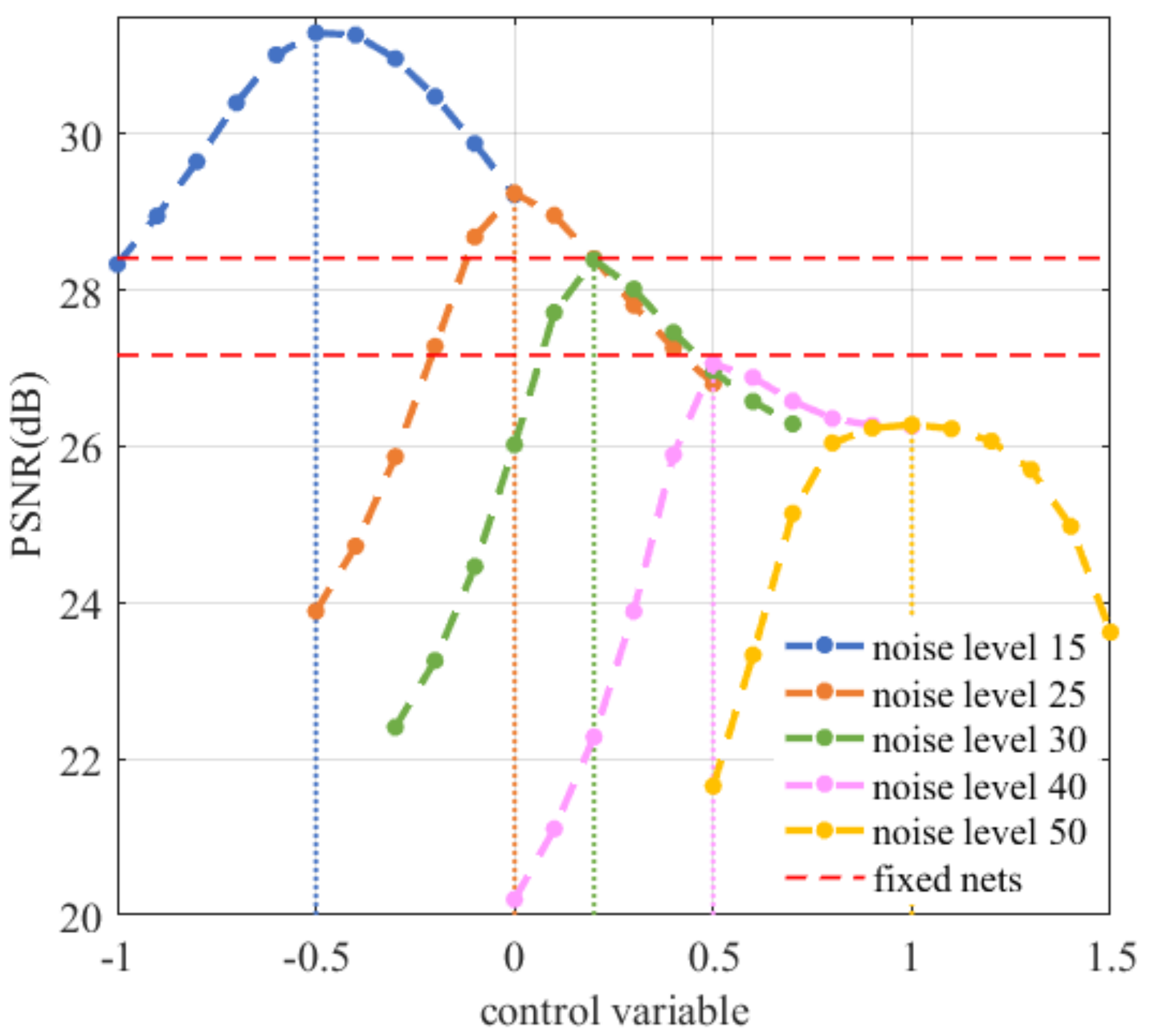}}
\subfigure[JPEG deblocking]{
\label{Ablation_alpha.sub.2}
\includegraphics[width=0.45\textwidth]{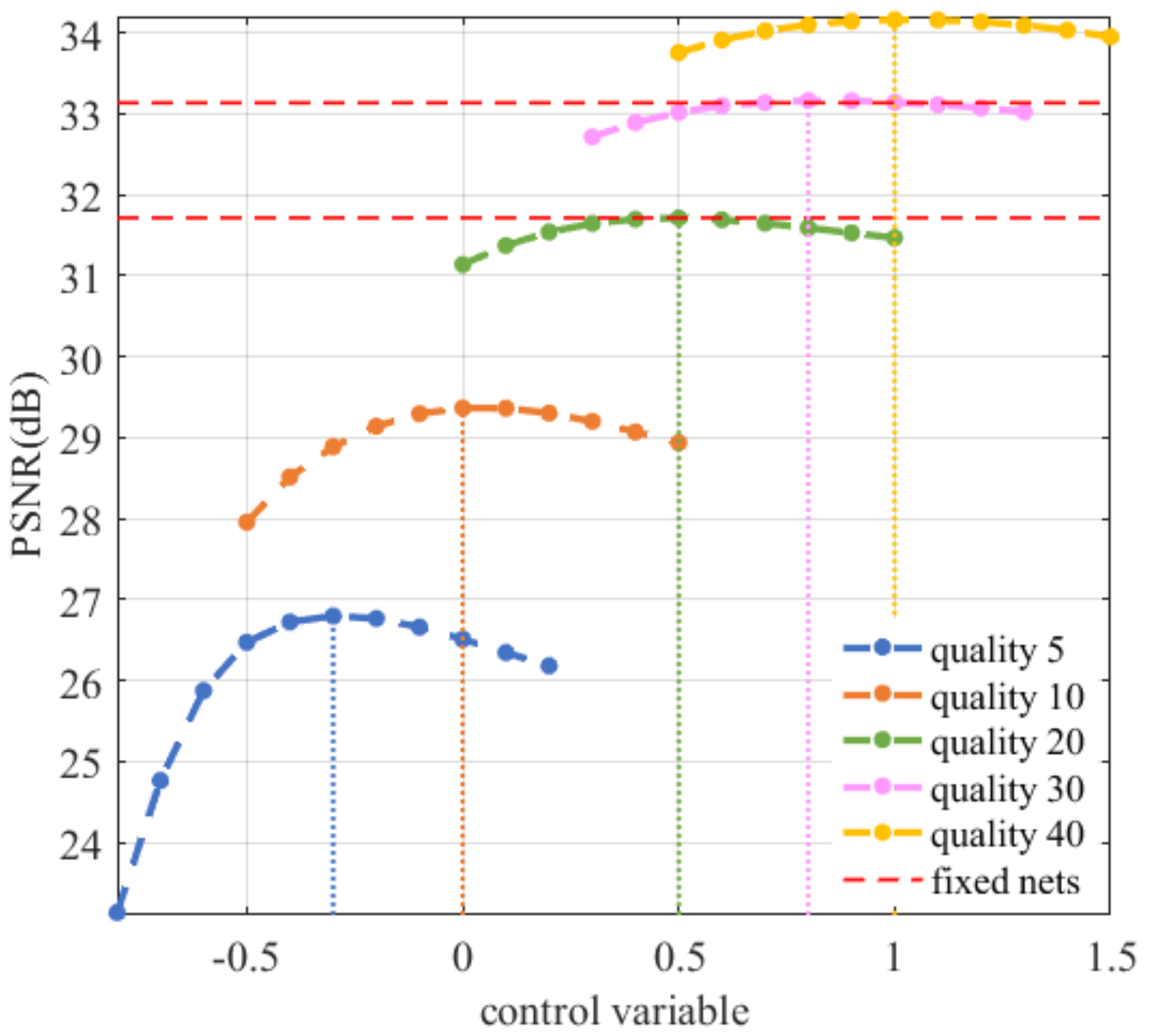}}
\caption{PSNR-$\alpha_{in}$ curve for five noise levels (\emph{i.e.}, 15, 25, 30, 40, and 50) on the (gray) BSD68 dataset and PSNR-$\alpha_{in}$ curve for five quality factors (\emph{i.e.}, 5, 10, 20, 30, and 40) on the LIVE1 dataset. Note that our CFSNet is only trained on two endpoints (\emph{e.g.}, noise level 25 and 50), but by adjusting the control variable $\alpha_{in}$, our CFSNet can still achieve similar performance compared with the fixed networks (red dotted line) at two intermediate working points (\emph{e.g.}, noise level 30 and 40).}
\label{psnr_alpha}
\end{figure}

\begin{bfseries} 
Control mechanism analyses
\end{bfseries}
In the ablation study of the main paper, we exhibit the reconstruction results of our CFSNet with different input control variables. Here, we provide more details on handling tasks with different degradation levels. To be more specific, we change control variable $\alpha_{in}$ with an interval of 0.1 for each degradation level. Besides, we use the main branch of our CFSNet as the fixed network, which is trained for each specific intermediate degradation level with corresponding data (red dotted line in the Fig.\ref{psnr_alpha}). As shown in Fig.\ref{psnr_alpha}, the observations can be summarized as: 1) Whether it is an image denoising task or a JPEG image deblocking task, we can always find the optimal control variable to obtain the result of minimal reconstruction distortion. On the other hand, the optimal $\alpha_{in}$ of different degradation levels is continuous. Taking image denoising as an example, the larger the noise level, the larger the optimal control variable. The noise level 15 even corresponds to a negative optimal control variable. This is because when we train our framework, we limit the two endpoint noise levels (noise level 25 and 50) to two endpoint control variables (0 and 1), respectively. Therefore, our framework establishes an implicit correspondence between control variables and degradation levels. 
2) Compared with the fixed networks (red dotted line), our CFSNet can achieve similar performance at two invisible intermediate working points of both restoration tasks, which indicates that our CFSNet can preferably emulate unknown working points. 
 In addition, we can empirically provide a reference value of the control variable for each degradation level to make our framework more user-friendly.

\begin{figure*}[!ht]
    \begin{minipage}[b]{.5\linewidth}
        \centering
        \includegraphics[width=0.96\linewidth]{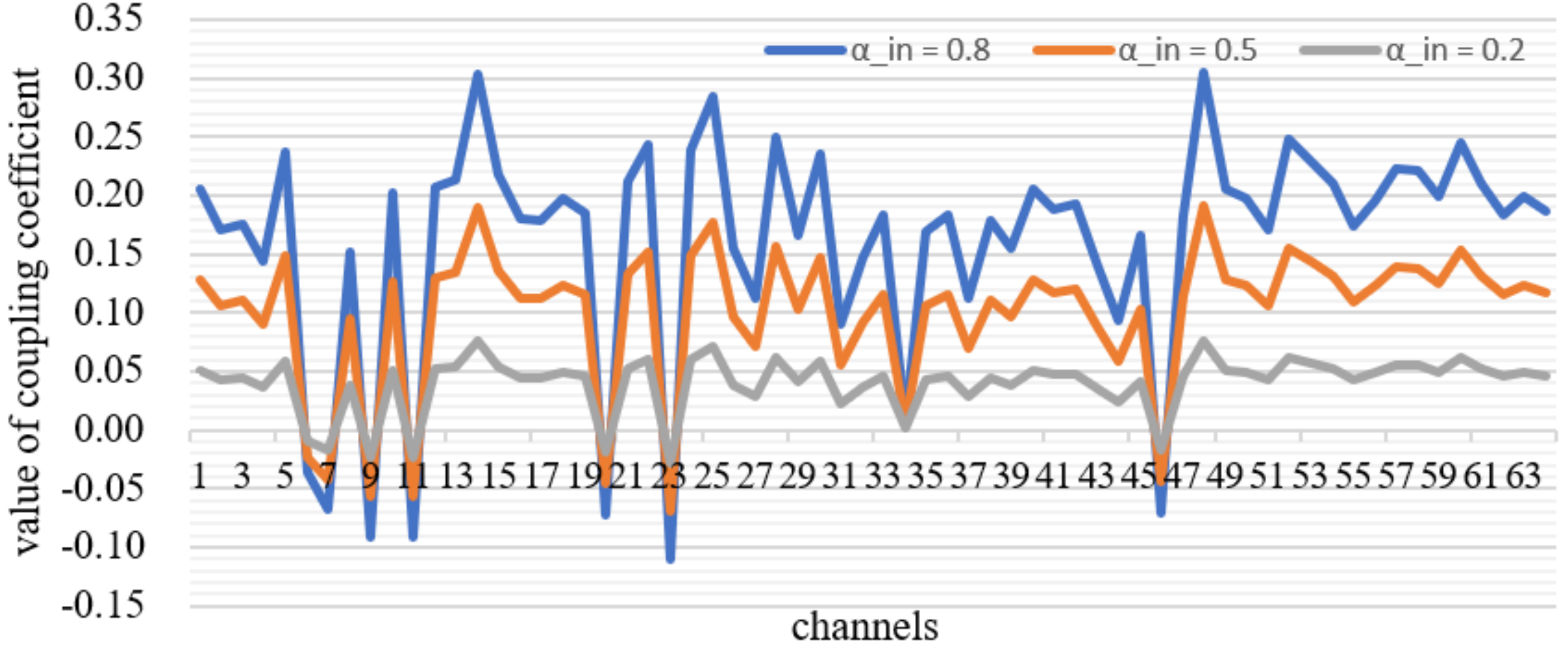}
        \makeatletter\def\@captype{figure}\makeatother\caption{The coupling coefficient curve of the second module in image denoising.}
        \label{alpha}
    \end{minipage}%
    \begin{minipage}[b]{.5\linewidth}
        \centering
        \begin{tabular}{|c|p{2.5cm}<{\centering}|p{2.5cm}<{\centering}|}
        \hline
        methods&Denoising &DeJPEG \\
        \hline
        \hline
        M10T5-top & 27.07($\alpha_{in}=0.6$)&33.10($\alpha_{in}=0.8$) \\
        \hline
        M10T5-last & 26.85($\alpha_{in}=0.5$) & 33.10($\alpha_{in}=0.9$) \\
        \hline
        M10T10 & 27.12($\alpha_{in}=0.5$) & 33.17($\alpha_{in}=0.8$) \\
        \hline
        M20T20 & 27.19($\alpha_{in}=0.5$) & 33.24($\alpha_{in}=0.8$) \\
        \hline
        M30T30 & 27.28($\alpha_{in}=0.5$) & 33.26($\alpha_{in}=0.8$) \\
        \hline
        \end{tabular}
        \makeatletter\def\@captype{table}\makeatother\caption{Model analysis. The average PSNR(dB) for unseen noise level $\sigma 40$ on (gray) BSD68 and unseen quality factor $q30$ on LIVE1. M and T denote the number of main blocks and the number of tuning blocks, respectively.}
        \label{model_analysis}
    \end{minipage}
\end{figure*}

\begin{figure*}[!ht]
\centering
\includegraphics[scale=0.335]{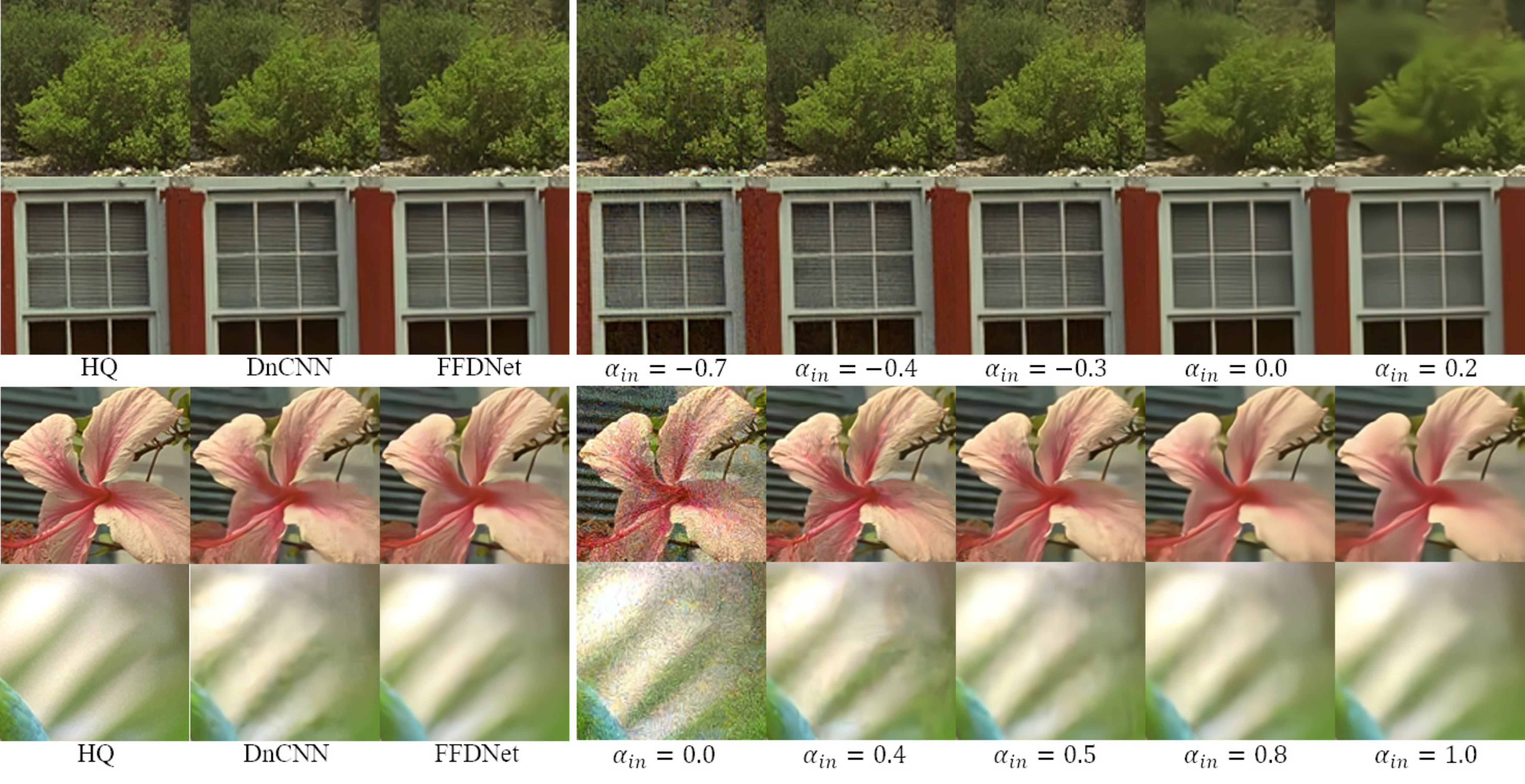}
\caption{Color image denoising results with unknown noise level $\sigma=15$ (first two rows) and $\sigma=40$ (last two rows). $\alpha_{in}=0.5$ and $\alpha_{in}=-0.3$ correspond to the highest PSNR results, respectively.}
\label{color_noise40}
\end{figure*}
\section{Model Analyses}
\label{Model_Analyses}

\begin{bfseries} 
The number of basic modules
\end{bfseries}
Here, we further evaluate the effect of the number of main blocks and tuning blocks. From the experimental results (Tab. \ref{model_analysis}), we observe that: 1) The more coupling modules, the better the reconstruction results. 2) Adding coupling modules only in the first 5 (M10T5-top) or last 5 (M10T5-last) layers is inferior to dense stacking mode (M10T10) in image denoising and JPEG image deblocking.

\begin{bfseries} 
The coupling coefficients learned
\end{bfseries}
First, since the manifold flatness of each layer of the network is different, the coefficients vary by channels and modules. This verifies that the coupling coefficients of CFSNet are learned adaptively from the training process. In addition, as shown in Fig. \ref{alpha}, the variation of coupling coefficients with different $\alpha_{in}$ is consistent, because each $\alpha_{in}$ corresponds to an intermediate latent representation.

\begin{figure*}[!ht]
\centering
\includegraphics[scale=0.37]{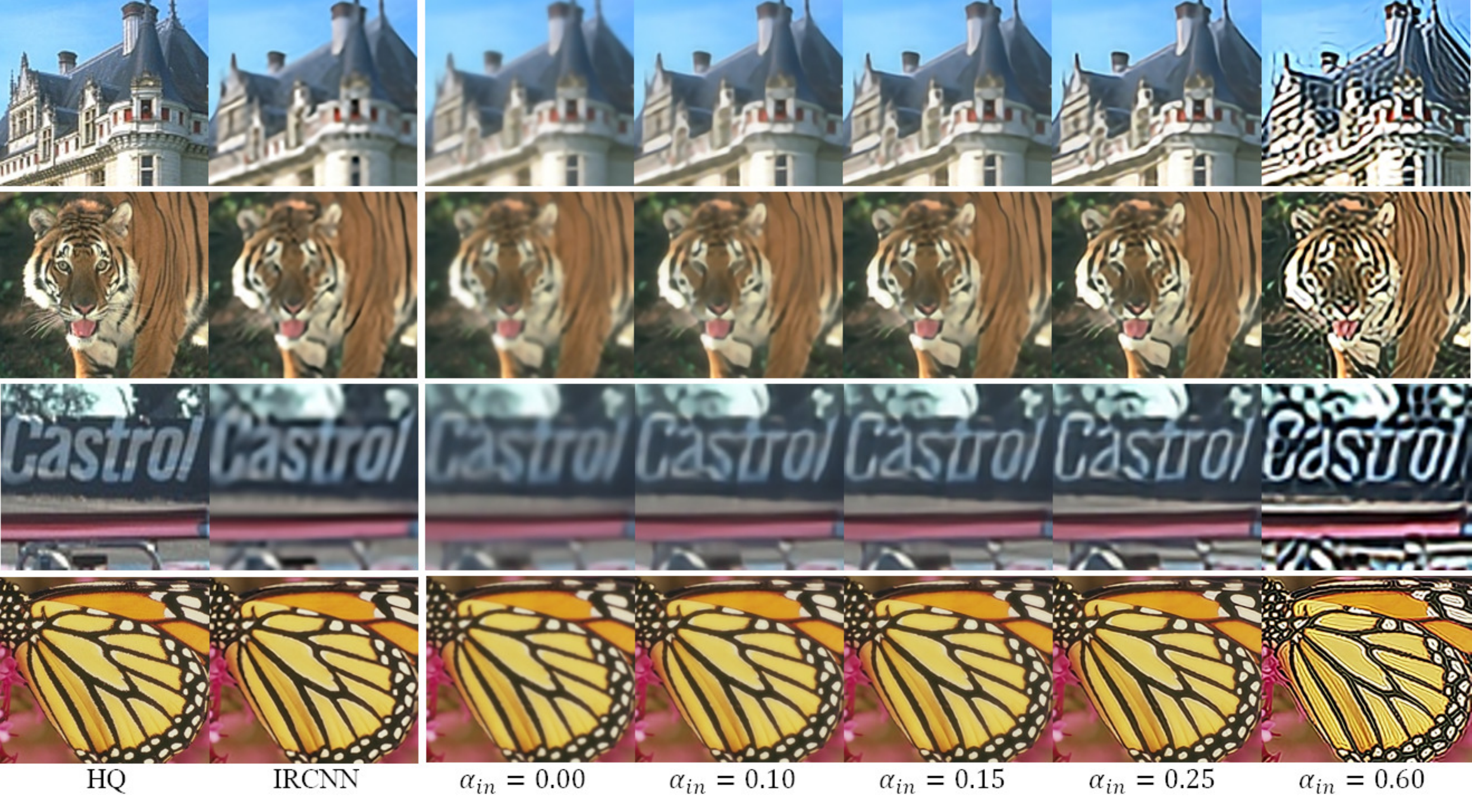}
\caption{Visual results of single image super-resolution with unseen degradation (the blur kernel is $7\times 7$ Gaussian kernel with standard deviation 1.6, the scale factor is 3). $\alpha_{in}=0.15$ corresponds to the highest PSNR results.}
\label{deblur_BD16}
\end{figure*}

\begin{figure*}[!ht]
\centering
\includegraphics[scale=0.37]{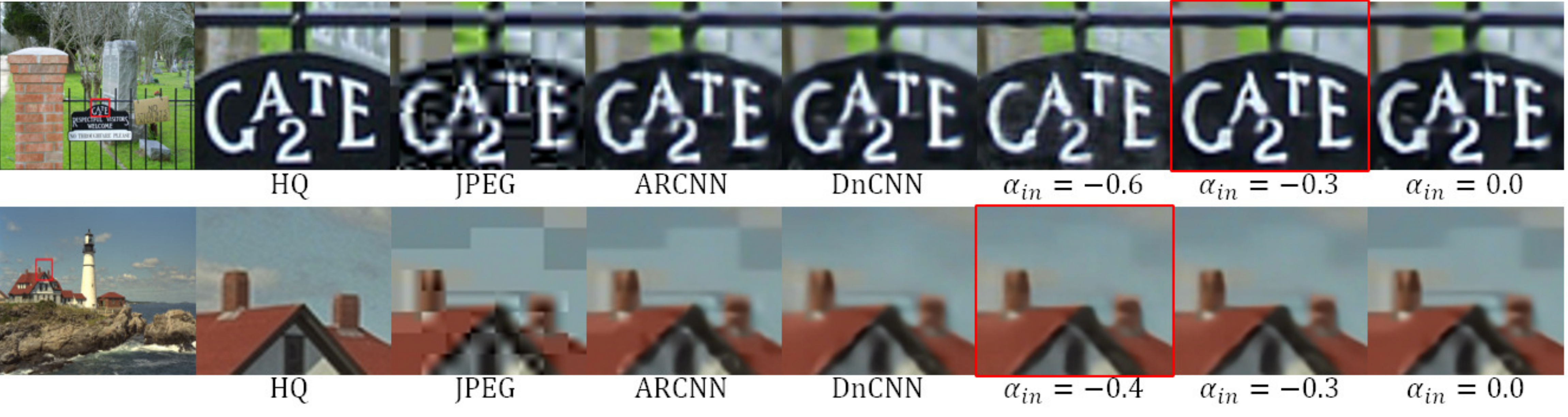}
\caption{JPEG image artifacts removal results of ``cemetry'' and ``lighthouse2'' (LIVE1) with unknown quality factor 5. $\alpha_{in}=-0.3$ corresponds to the highest PSNR results, and the best visual results are marked with red boxes.}
\label{jpeg_20}
\end{figure*}

\subsection{Color Image Denoising}
\label{Color_Image_Denoising}

In the main paper, we show the superiority of our CFSNet in gray image denoising task. Here, we further extend our model to color image denoising. We generate noisy color images by adding AWGN noise to clean RGB images with different noise levels $\sigma = 15, 25, 30, 40,$ and $50$. We evaluate our CFSNet using the Kodak24 \cite{franzen1999kodak} dataset. For the specific implementation, we use the same model settings as the gray image denoising task.

We show the color image denoising visual results in Fig. \ref{color_noise40}. We can see that: 1) Our CFSNet can play a role in controlling the trade-off between noise reduction and detail preservation. In particular, compared with DnCNN-B \cite{zhang2017beyond} and FFDNet \cite{zhang2018ffdnet}, our CFSNet can even achieve similar visual effects when dealing with the noise level $\sigma=15$ which beyond the preset range ($\left [ 25, 50 \right ]$). This further verifies the rationality of our theoretical analysis. 2) The optimal visual quality is image specific or scene specific. For example, ``flower'' (3rd row) enjoys more realistic texture when $\alpha_{in}=0.4$, while ``leaf'' (4th row) enjoys smoother artifact-free result when $\alpha_{in}=1.0$. However, the fixed method (\emph{e.g.}, DnCNN-B \cite{zhang2017beyond}) can not adequately meet specific needs. Therefore, it is necessary to adjust the reconstruction results according to specific goals.


\subsection{General Image Super-resolution}
\label{SR_blur}
In the main paper, we show the perception-distortion trade-off of image super-resolution modeled by a simple bicubic downsampling operation. Here, in order to further demonstrate the flexibility of our CFSNet, we extend our framework to a more challenging degradation model. To be specific, when we train the main branch in Step 1, we still adopt bicubic downsampling as the degradation setting. However, when we train the tuning branch in Step 2, we first blur the HR image by Gaussian kernel of size $17\times 17$ with standard deviation 2.6, then we bicubic downsample it with scale factor 3 to produce an LR image. For testing, we use three
standard benchmark datasets: Set5 \cite{bevilacqua2012low}, B100 \cite{martin2001database}, Urban100 \cite{huang2015single}, and we follow \cite{zhang2017learning} to use a popular $7\times 7$ Gaussian kernel with width 1.6 which never appear in training process. Besides, both branches of our framework are trained based on the same MAE loss. In more details, we adopt a small model that contains 10 main blocks and 10 tuning blocks (i.e., $M=10$) for a fair comparison.

Fig. \ref{deblur_BD16} shows the visual results of the proposed CFSNet. Tab. \ref{BD_SR} shows the average PSNR results for SRCNN \cite{dong2016image}, VDSR \cite{kim2016accurate}, IRCNN \cite{zhang2017learning} and our CFSNet. Several observations can be summarized as follows: 1) Our CFSNet can make a nice compromise between blur removal and detail sharpening. Specifically, the large control variable $\alpha_{in}$ leads to over sharpening artifacts. In contrast, incomplete blur elimination results can be observed using a small control variable $\alpha_{in}$. 2) Our CFSNet achieves the best PSNR results on all test datasets. 3) the lowest distortion result ($\alpha_{in}=0.15$) does not necessarily mean the best visual effects. For example, the result of ``tiger'' is more visually acceptable when $\alpha_{in}=0.25$.

\begin{table}[!ht]
\centering
\caption{The average general image super-resolution results of PSNR (dB) on three benchmark datasets. Note that the degradation settings of all the testing images are not included in the training stages of our CFSNet.}
\begin{tabular}{|c|c|c|c|c|}
\hline
Dataset&SRCNN&VDSR&IRCNN&CFSNet\\
\hline
\hline
Set5&32.05 & 33.25 &33.38&33.50\\
\hline
B100&28.13& 28.57 &28.65&28.79\\
\hline
Urban100&25.70&26.61&26.77&27.33\\
\hline
\end{tabular}
\label{BD_SR}
\end{table}

\begin{figure*}[!ht]
\centering
\includegraphics[width=17.4cm, height=10cm]{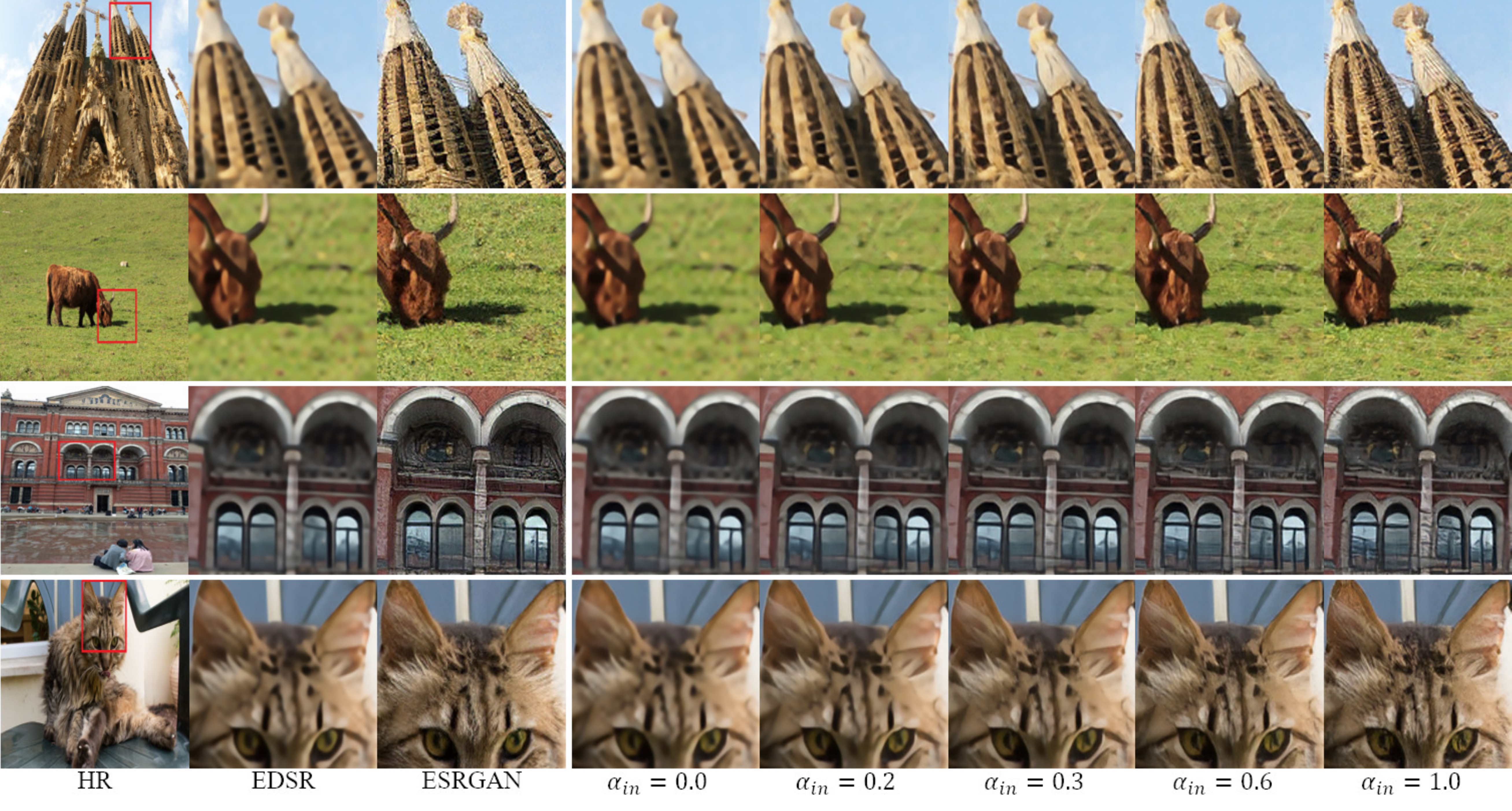}
\caption{Perceptual and distortion balance for $4\times$ image super-resolution.}
\label{SR}
\end{figure*}

\subsection{Additional Results}
\label{Additional_Results}
We present additional perceptual and distortion balance results for $4\times$ image super-resolution in Fig. \ref{SR}, and additional JPEG image artifacts removal results in Fig. \ref{jpeg_20}.

\end{document}